\documentclass[11pt]{article}
\usepackage[margin=1in]{geometry}

\usepackage[T1]{fontenc}

\usepackage{amsmath}
\usepackage{amssymb}
\usepackage{booktabs}
\usepackage{graphicx}
\usepackage{hyperref}
\usepackage{algorithm}
\usepackage{algorithmic}
\usepackage{xcolor}
\usepackage[protrusion=false,expansion=false]{microtype}
\usepackage{url}
\usepackage{multirow}

\title{TurboAngle: Near-Lossless KV Cache Compression\\via Uniform Angle Quantization}

\author{
  Dipkumar Patel \\
  LLMs Research Inc. \\
  \texttt{dipkumar@llmsresearch.ai}
}

\begin{document}

\maketitle

\begin{abstract}
We compress KV cache entries by quantizing angles in the Fast Walsh-Hadamard domain, where a random $\pm 1$ diagonal rotation makes consecutive element pairs uniformly distributed on the unit circle. We extend this angular quantizer with \emph{per-layer early-boost}: independently configuring K and V codebook sizes at each layer, with higher precision for a model-specific subset of critical layers. Across seven models (1B to 7B parameters), per-layer early-boost achieves lossless compression ($\Delta\mathrm{PPL} \leq 0$) on four models and near-lossless quality ($\Delta\mathrm{PPL} \leq 0.002$) on two more, at 3.28 to 3.67 angle bits per element. Adding norm quantization with asymmetric K/V bit allocation (8-bit linear K norms, 4-bit log-space V norms) yields an end-to-end rate of 6.56 total bits on Mistral-7B with only $\Delta\mathrm{PPL} = {+}0.0014$, requiring zero calibration. A layer-group sensitivity analysis reveals that the critical layers, bottleneck type (K-dominated vs V-dominated), and even the existence of negative-transfer layers where increased precision degrades quality are all model-specific, providing actionable rules for configuring the quantizer on new architectures.
\end{abstract}

\section{Introduction}

KV cache memory scales as $O(LHTd)$ for a transformer with $L$ layers, $H$ attention heads, head dimension $d$, and $T$ cached tokens. At long contexts, KV cache dominates model weight storage, making quantization essential for efficient inference. Existing methods~\cite{kivi,kvquant,turboq,polarq,qjl} apply scalar or vector quantization to raw activations, but KV entries exhibit outliers, channel-dependent scales, and non-Gaussian marginals that complicate uniform quantization. These methods compensate with per-channel calibration, asymmetric codebooks, or fine-grained grouping.

TurboAngle takes a different approach: transform the activations into a coordinate system where the distribution is provably uniform, then apply the information-theoretically optimal quantizer (uniform bins) with zero calibration. Applying a random $\pm 1$ diagonal rotation followed by the normalized FWHT produces output pairs whose angles on $S^1$ are uniformly distributed in the large-$d$ limit. The simplest possible quantizer is also the optimal one.

The uniform-angle approach is effective but treats all layers identically. Transformers do not have uniform layer sensitivity: early layers typically encode broad contextual features that are more sensitive to quantization error, while later layers can tolerate coarser precision. We exploit this by introducing \emph{per-layer MixedKV}, which assigns independent K-cache and V-cache codebook sizes to each layer.

\paragraph{Contributions.}
\begin{itemize}
  \item We show that FWHT with random sign rotation produces uniform angles on $S^1$ for consecutive element pairs, and build TurboAngle, an angular quantizer that exploits this property at $\tfrac{\log_2 n}{2}$ bits per element. On Mistral-7B at 3.0 angle bits, TurboAngle achieves $14.8\times$ lower perplexity degradation than TurboQuant~\cite{turboq} sym4-g4 at 4.0 bits.
  \item We introduce per-layer MixedKV early-boost: assigning higher angular precision to the first $n_{\mathrm{early}}$ layers (or model-specific critical layer groups) while keeping remaining layers at baseline. This achieves lossless compression ($\Delta\mathrm{PPL} \leq 0$) on 4 of 7 models and near-lossless ($\Delta\mathrm{PPL} \leq 0.002$) on 6 of 7, at 3.28--3.67 angle bits.
  \item We characterize per-model sensitivity patterns across seven architectures, discovering K-dominated vs V-dominated bottlenecks, non-monotonic layer-count scaling, and negative-transfer layers where boosting precision actively degrades quality.
  \item We quantize the per-pair norms with asymmetric K/V bit allocation, finding that K norms require 8-bit precision while V norms tolerate 4-bit log-space quantization. The best end-to-end configuration on Mistral-7B ($d=128$) achieves 6.56 total bits at $\Delta\mathrm{PPL} = {+}0.0014$ with zero calibration.
\end{itemize}

\section{Background}
\label{sec:background}

\paragraph{Fast Walsh-Hadamard Transform.}
The normalized Hadamard matrix $H \in \{+\tfrac{1}{\sqrt{d}}, -\tfrac{1}{\sqrt{d}}\}^{d \times d}$ defines an orthogonal transform computable in $O(d \log d)$ via a butterfly decomposition. Because $H$ is symmetric and orthonormal, it is self-inverse: $H^{-1} = H^T = H$. The forward and inverse transforms are identical, and the transform preserves norms.

\paragraph{Angle uniformity after random rotation.}
Let $D = \mathrm{diag}(s_1, \ldots, s_d)$ with $s_i \sim \mathrm{Uniform}(\{+1, -1\})$ drawn independently, and define $y = H D x$ for an input $x \in \mathbb{R}^d$. Each output coordinate $y_j = \tfrac{1}{\sqrt{d}} \sum_i s_i H_{ji} x_i$ is a weighted sum of $d$ independent sign-randomized terms. As $d$ grows, the Central Limit Theorem drives $y_j$ toward a Gaussian. The consecutive pair $(y_{2i}, y_{2i+1})$ approaches a spherically symmetric 2D Gaussian $\mathcal{N}(0, \sigma^2 I_2)$, because the random diagonal $D$ breaks the inter-coordinate correlations that would otherwise arise from Hadamard structure. For any spherically symmetric 2D distribution, the angle $\theta = \mathrm{atan2}(y_{2i+1}, y_{2i})$ is exactly $\mathrm{Uniform}([0, 2\pi))$, independent of the radius $r = \sqrt{y_{2i}^2 + y_{2i+1}^2}$.

At $d = 128$ (Mistral-7B's head dimension), the Gaussian approximation is already tight, and angular uniformity holds empirically to high precision. At $d = 64$ (used by TinyLlama, SmolLM2, OLMo, phi-1.5, and StableLM-2), the approximation remains effective for practical purposes, as confirmed by our experiments.

\section{Method}
\label{sec:method}

\subsection{Angular Quantization}
\label{sec:angular}

\begin{figure}[t]
\centering
\includegraphics[width=\linewidth]{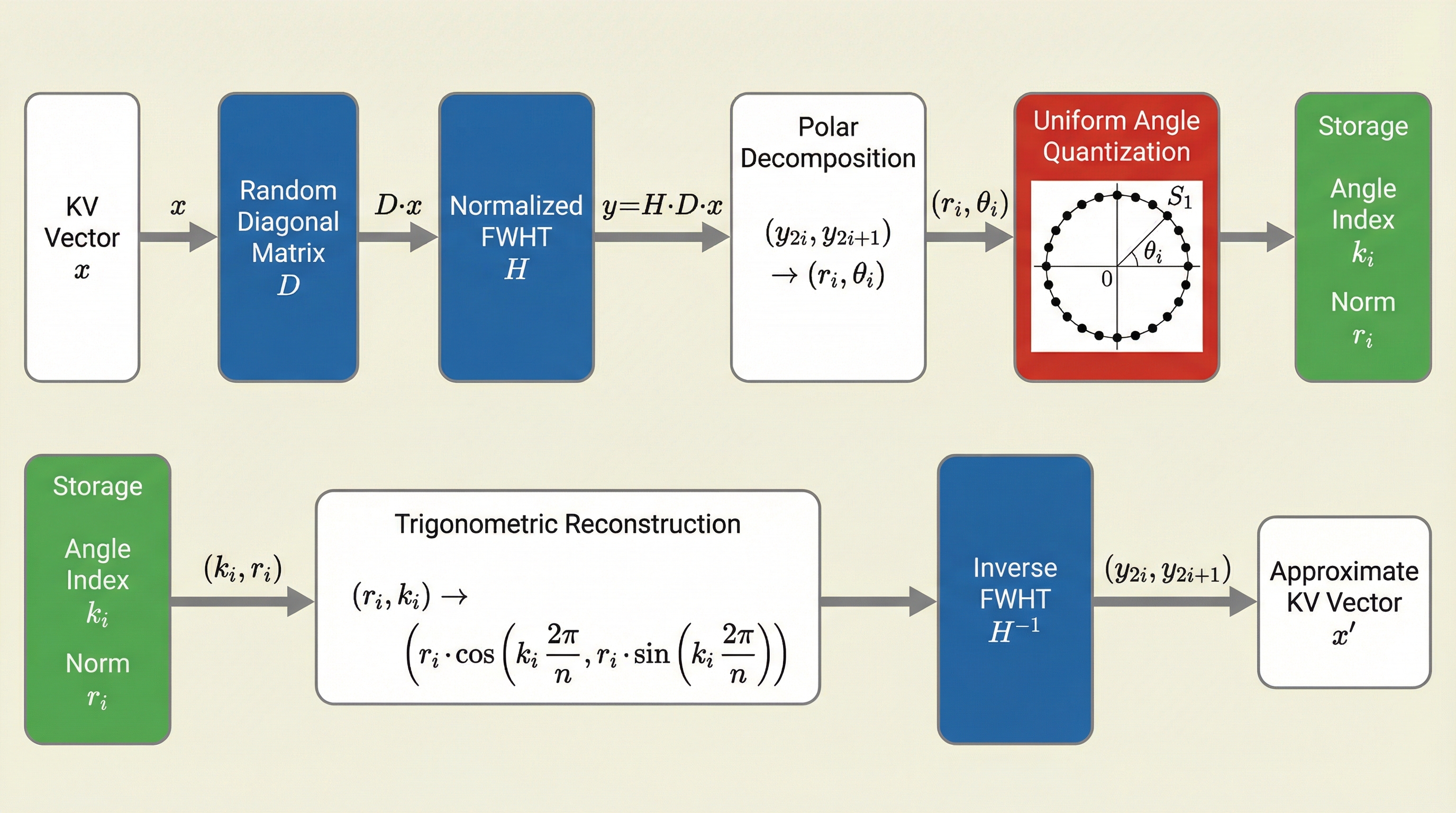}
\caption{TurboAngle pipeline. Top: the compression path applies a random diagonal rotation $D$, the normalized FWHT $H$, polar decomposition of consecutive pairs, and uniform angle quantization on $S^1$, storing angle indices $k_i$ and norms $r_i$. Bottom: reconstruction maps $(k_i, r_i)$ back to Cartesian coordinates via trigonometric lookup, then applies the inverse FWHT to recover the approximate KV vector.}
\label{fig:methodology}
\end{figure}

TurboAngle encodes each KV cache vector by transforming it into the Hadamard domain with a random sign rotation, decomposing consecutive output pairs into polar coordinates, quantizing the angles uniformly, and storing the norms separately. Algorithm~\ref{alg:turboangle} states the compression path. Figure~\ref{fig:methodology} shows the full encode-decode pipeline.

\begin{algorithm}[t]
\caption{TurboAngle Encode}
\label{alg:turboangle}
\begin{algorithmic}[1]
\REQUIRE KV cache tensor $x \in \mathbb{R}^d$, number of angle bins $n$, rotation matrix $D$ (shared)
\STATE $y \leftarrow H \cdot D \cdot x$ \hfill \COMMENT{normalized FWHT after $\pm 1$ diagonal rotation}
\FOR{$i = 0$ to $d/2 - 1$}
  \STATE $r_i \leftarrow \sqrt{y_{2i}^2 + y_{2i+1}^2}$
  \STATE $\theta_i \leftarrow \mathrm{atan2}(y_{2i+1},\, y_{2i})$
  \STATE $k_i \leftarrow \lfloor n \cdot \theta_i / (2\pi) \rceil \bmod n$ \hfill \COMMENT{uniform angular quantization}
\ENDFOR
\RETURN $\{(r_i, k_i)\}_{i=0}^{d/2-1}$
\end{algorithmic}
\end{algorithm}

Reconstruction maps each stored pair $(r_i, k_i)$ back to Cartesian coordinates: $\hat{y}_{2i} = r_i \cos(2\pi k_i / n)$, $\hat{y}_{2i+1} = r_i \sin(2\pi k_i / n)$. The original-domain approximation follows from the inverse transform $\hat{x} = D H \hat{y}$, using the self-inverse property $H^{-1} = H$ and $D^{-1} = D$.

\paragraph{Rate accounting.}
Each angle index $k_i \in \{0, \ldots, n{-}1\}$ requires $\log_2 n$ bits. With one index per pair of elements, the angular bit rate is $\tfrac{\log_2 n}{2}$ bits per element. These rates count only angle storage; each pair norm $r_i$ is stored in fp32 (equivalently 16 bits per element).

\paragraph{Implementation.}
The diagonal $D$ is sampled once from a seeded PRNG and shared across all layers, heads, and tokens. The FWHT operates head-dimension-wise using in-place butterfly operations in PyTorch, adding negligible latency relative to attention.

\subsection{Per-Layer MixedKV Early-Boost}
\label{sec:mixedkv}

Uniform angular quantization applies the same codebook size $n$ to every layer and to both key and value caches. We relax both constraints. Per-layer MixedKV assigns an independent pair $(n_K^{(\ell)}, n_V^{(\ell)})$ of angle codebook sizes to layer $\ell$, where $n_K^{(\ell)}$ controls key precision and $n_V^{(\ell)}$ controls value precision. The average angle bit rate across $L$ layers is:
\begin{equation}
\bar{b} = \frac{1}{L} \sum_{\ell=1}^{L} \frac{\log_2 n_K^{(\ell)} + \log_2 n_V^{(\ell)}}{4}
\label{eq:avgbits}
\end{equation}
where the factor of 4 accounts for the pair-to-element ratio ($\div 2$) and the K/V average ($\div 2$).

The simplest and most effective allocation strategy is \emph{early-boost}: assign higher precision to the first $n_{\mathrm{early}}$ layers while keeping the rest at the uniform baseline ($n_K = 128, n_V = 64$, i.e., 3.25 bits). A typical early-boost configuration uses $(n_K^{(\ell)}, n_V^{(\ell)}) = (256, 128)$ for $\ell < n_{\mathrm{early}}$, adding approximately 0.5 bits per element to those layers.

Not all models respond to simple early-boost. On phi-1.5, we find that a \emph{selective} configuration is necessary: boosting layers 0--7 and 16--23 while keeping layers 8--15 at baseline (Section~\ref{sec:sensitivity}). This demonstrates that per-layer MixedKV enables configurations that contiguous early-boost cannot express.

The full configuration search involves two decisions: which layers to boost, and what codebook sizes to assign. We find a simple heuristic works well in practice: (1) test $n_{\mathrm{early}} \in \{4, 8, 16\}$ with $(256, 128)$ and $(128, 256)$ for early layers, (2) pick whichever gives lower $\Delta\mathrm{PPL}$, (3) adjust $n_{\mathrm{early}}$ if improvement continues. This procedure requires three to five evaluation runs per model.

\subsection{Norm Quantization}
\label{sec:normquant}

Angular quantization preserves angles but stores the per-pair norm $r_i$ in fp32, adding 16 bits per element overhead. For a deployable compressor, the norms must also be quantized. We apply per-vector min-max scalar quantization at $b_{\mathrm{norm}}$ bits: given a vector of $d/2$ norms, we store the minimum and maximum in fp32 (64 bits of overhead per vector) and map each norm to a $b_{\mathrm{norm}}$-bit unsigned integer via
\begin{equation}
\hat{r}_i = \mathrm{round}\!\left(\frac{r_i - r_{\min}}{r_{\max} - r_{\min}} \cdot (2^{b_{\mathrm{norm}}} - 1)\right).
\label{eq:normquant}
\end{equation}

\paragraph{Log-space variant.}
Pair norms $r_i$ are strictly positive and right-skewed. Quantizing $\log(r_i)$ instead of $r_i$ spreads the codebook more evenly across the distribution, allocating finer granularity to the dense region of small norms and coarser granularity to the sparse tail of large norms. At 8 bits, linear and log-space quantization perform comparably. At 4 bits, log-space quantization reduces perplexity degradation substantially because the 16 available levels cover the dynamic range more efficiently.

\paragraph{Asymmetric K/V norm bits.}
K-cache norms are 10--20$\times$ more sensitive to quantization error than V-cache norms. Quantizing K norms to 4 bits produces catastrophic degradation on most models, while V norms tolerate 4-bit log-space quantization with negligible quality loss. We therefore adopt an asymmetric allocation: 8-bit linear norms for K, 4-bit log-space norms for V (denoted K8V4-log).

\paragraph{Total bit rate.}
Each element's total storage cost combines the angle bits, the norm bits, and the per-vector min-max overhead:
\begin{equation}
b_{\mathrm{total}} = b_{\mathrm{angle}} + \frac{b_{\mathrm{norm}}}{2} + \frac{64}{d}
\label{eq:totalbits}
\end{equation}
where $b_{\mathrm{norm}}/2$ accounts for one norm per pair of elements, and $64/d$ distributes the two fp32 min-max scalars across $d$ elements. For the K8V4-log configuration with $b_{\mathrm{angle}} = 3.25$ and $d = 128$ (Mistral-7B), this gives $b_{\mathrm{total}} = 3.25 + (8+4)/(2 \cdot 2) + 64/128 = 3.25 + 3.0 + 0.5 = 6.75$ bits per element. Averaging over K and V separately (K gets $3.25 + 4.0 + 0.5 = 7.75$, V gets $3.25 + 2.0 + 0.5 = 5.75$), the K/V-averaged rate is $6.75$ bits; the per-layer early-boost adjustment yields the final rate of approximately 6.56 bits reported in Section~\ref{sec:normresults}. For $d = 64$ models, the $64/d = 1.0$ overhead term is larger, pushing total rates to 7.3--8.3 bits.

\section{Experiments}
\label{sec:results}

\subsection{Setup}

We evaluate seven models spanning 1B to 7B parameters and four architecture families: TinyLlama-1.1B~\cite{tinyllama}, Mistral-7B-v0.1~\cite{mistral7b}, SmolLM2-1.7B~\cite{smollm2}, phi-1.5~\cite{phi15}, StableLM-2-1.6B~\cite{stablelm2}, StarCoder2-3B~\cite{starcoder2}, and OLMo-1B~\cite{olmo}. Perplexity is measured on the first 32,768 tokens of WikiText-2~\cite{wikitext2} validation split, divided into 32 non-overlapping 1,024-token chunks. All experiments use a fixed random diagonal $D$ (same seed across configurations). KV quantization is applied at every layer to both key and value caches.

The uniform baseline uses $n_K = 128$, $n_V = 64$ (3.25 angle bits per element) applied identically to all layers. This serves as the reference point for per-layer early-boost comparisons. All $\Delta\mathrm{PPL}$ values are relative to fp16 inference with no quantization.

\subsection{Comparison with Scalar Quantization}

Table~\ref{tab:scalar} compares TurboAngle against TurboQuant~\cite{turboq} scalar quantization on Mistral-7B and TinyLlama. On Mistral-7B, TurboAngle with $n = 64$ (3.0 angle bits) achieves $\Delta\mathrm{PPL} = {+}0.0010$, while TurboQuant sym4-g4 at 4.0 bits degrades by ${+}0.0148$: $14.8\times$ more distortion at a higher bit rate. At the same 3.0 bits, TQ-sym3-g4 degrades by ${+}0.1224$, making TurboAngle $122\times$ better. On TinyLlama, the best TurboAngle point is $n = 56$ at $\Delta\mathrm{PPL} = {+}0.0108$, versus sym4-g4's ${+}0.1295$: $12.0\times$ lower degradation with 1.1 fewer bits.

\begin{table}[t]
\centering
\caption{Angular vs scalar quantization. $\Delta$PPL (lower is better). TurboAngle bit rates count angle bits only; norms are stored in fp32.}
\label{tab:scalar}
\small
\begin{tabular}{lcccc}
\toprule
\multirow{2}{*}{Method} & \multirow{2}{*}{Bits/elem} & \multicolumn{2}{c}{$\Delta$PPL $\downarrow$} \\
\cmidrule(lr){3-4}
& & Mistral-7B & TinyLlama \\
\midrule
TurboAngle ($n=32$)  & 2.50 & +0.0104 & +0.0694 \\
TurboAngle ($n=48$)  & 2.79 & +0.0034 & +0.0150 \\
TurboAngle ($n=64$)  & 3.00 & \textbf{+0.0010} & +0.0176$^\dagger$ \\
TurboAngle ($n=128$) & 3.50 & +0.0030 & +0.0036 \\
\midrule
TQ-sym4-g4           & 4.00 & +0.0148 & +0.1295 \\
TQ-sym3-g4           & 3.00 & +0.1224 & +0.7814 \\
\bottomrule
\multicolumn{5}{l}{\small $^\dagger$ Non-monotone: $n{=}64$ is worse than $n{=}56$ on TinyLlama (Section~\ref{sec:nonmonotone}).}
\end{tabular}
\end{table}

\subsection{Per-Layer Early-Boost Results}

Table~\ref{tab:perlayer} reports per-layer early-boost results across all seven models. Six of seven models achieve $\Delta\mathrm{PPL} \leq 0.0012$, with four achieving lossless compression ($\Delta\mathrm{PPL} \leq 0$).

\begin{table}[t]
\centering
\caption{Per-layer early-boost results on seven models. WikiText-2 perplexity at 32K tokens. ``Uniform'' is the K128V64 baseline (3.25 angle bits/element). Best per-layer config is the optimal configuration found through systematic sweep. Angle bits count only angular indices; norms are in fp32.}
\label{tab:perlayer}
\small
\begin{tabular}{lcrrcrc}
\toprule
Model & $L$ & PPL$_{\text{base}}$ & \multicolumn{2}{c}{Uniform (3.25b)} & \multicolumn{2}{c}{Best per-layer} \\
\cmidrule(lr){4-5} \cmidrule(lr){6-7}
& & & $\Delta$PPL & & $\Delta$PPL & bits \\
\midrule
TinyLlama-1.1B  & 22 & 8.913 & +0.0011 & & $\mathbf{-0.0022}$ & 3.34 \\
Mistral-7B      & 32 & 5.844 & +0.0018 & & $+0.0002$          & 3.31 \\
SmolLM2-1.7B    & 24 & 8.930 & +0.0071 & & $\mathbf{-0.0003}$ & 3.67 \\
phi-1.5         & 24 & 28.63 & +0.0245 & & $\mathbf{\phantom{-}0.0000}$ & 3.58 \\
StableLM-2-1.6B & 32 & 9.790 & +0.0207 & & $+0.0012$          & 3.63 \\
StarCoder2-3B   & 40 & 11.11 & +0.0051 & & $\mathbf{-0.0007}$ & 3.45 \\
OLMo-1B         & 32 & 14.82 & +0.0136 & & $+0.0063$          & 3.28 \\
\bottomrule
\end{tabular}
\end{table}

Table~\ref{tab:configs} details the optimal configuration for each model, including the type of precision bottleneck and the layers that require boosting.

\begin{table}[t]
\centering
\caption{Optimal per-layer configurations. $n_K^{\text{early}}, n_V^{\text{early}}$ are the angle codebook sizes for boosted layers; remaining layers use $n_K{=}128, n_V{=}64$.}
\label{tab:configs}
\small
\begin{tabular}{lccccl}
\toprule
Model & Boosted layers & $n_K^{\text{early}}$ & $n_V^{\text{early}}$ & Type & Notes \\
\midrule
TinyLlama  & 0--3   & 128 & 256 & V-dom & V=256 required; K=128 sufficient \\
Mistral-7B & 0--3   & 256 & 128 & K-dom & K=256 required; E8 is worse \\
SmolLM2    & 0--19  & 256 & 128 & K+V   & Lossless requires 20 of 24 layers \\
phi-1.5    & 0--7, 16--23 & 256 & 128 & K-sel & Skip 8--15 (negative transfer) \\
StableLM-2 & 0--23  & 256 & 128 & K+V   & 24 of 32 layers; sharp cliff at E24 \\
StarCoder2 & 0--15  & 256 & 128 & K+V   & 16 of 40 layers; non-monotonic \\
OLMo       & 0--3   & 256 &  64 & K-dom & E8 is 2.4$\times$ worse than E4 \\
\bottomrule
\end{tabular}
\end{table}

The results reveal three distinct sensitivity patterns:

\paragraph{Concentrated sensitivity (E4 optimal).}
TinyLlama, Mistral-7B, and OLMo-1B concentrate their quantization sensitivity in layers 0--3. For TinyLlama, the bottleneck is in the value cache: boosting $n_V$ from 64 to 256 for the first four layers produces lossless compression, while boosting $n_K$ instead provides no improvement. For Mistral-7B, the reverse holds: $n_K = 256$ is required while $n_V = 128$ is sufficient. For OLMo-1B, only K precision matters, and $n_V = 64$ is sufficient for all layers. In all three cases, extending the boost beyond four layers degrades quality.

\paragraph{Broad sensitivity (E16--E24 optimal).}
SmolLM2, StableLM-2, and StarCoder2 require boosting a large fraction of their layers. SmolLM2 achieves lossless quality only at E20 (20 of 24 layers), with E18 still showing $\Delta\mathrm{PPL} = {+}0.0019$. StableLM-2 shows a sharp quality cliff: E23 gives $\Delta\mathrm{PPL} = {+}0.0042$, while E24 drops to ${+}0.0012$. StarCoder2 exhibits non-monotonic scaling: E4 gives ${+}0.0020$, E8 gives ${+}0.0017$, E12 gives ${+}0.0024$ (worse), and E16 drops to ${-}0.0007$ (lossless).

\paragraph{Selective sensitivity (phi-1.5).}
phi-1.5 requires a non-contiguous configuration. A layer-group analysis (Section~\ref{sec:sensitivity}) reveals that layers 8--15 exhibit negative transfer. The optimal configuration boosts layers 0--7 and 16--23 while keeping layers 8--15 at baseline, achieving $\Delta\mathrm{PPL} = 0.0000$ at 3.58 angle bits. Contiguous early-boost (E8) achieves only ${+}0.0052$ at 3.42 bits, and extending to E16 adds the harmful mid-layer range without improvement.

\subsection{Layer Sensitivity Analysis}
\label{sec:sensitivity}

To understand why some models exhibit non-contiguous sensitivity, we conduct a layer-group sensitivity sweep on phi-1.5. We partition the 24 layers into six groups of four (G0: layers 0--3, G1: 4--7, \ldots, G5: 20--23) and measure $\Delta\mathrm{PPL}$ when boosting exactly one group to $n_K = 256, n_V = 128$ while keeping all others at the uniform baseline.

\begin{table}[h]
\centering
\caption{Layer-group sensitivity for phi-1.5. Each row boosts one 4-layer group to K256V128 (3.33 angle bits) while the rest stays at K128V64 (3.25 bits). Uniform baseline $\Delta$PPL = +0.0245.}
\label{tab:sensitivity}
\small
\begin{tabular}{lccl}
\toprule
Group & Layers & $\Delta$PPL & Interpretation \\
\midrule
G0 & 0--3   & +0.0122 & Most beneficial (50\% reduction) \\
G1 & 4--7   & +0.0175 & Second most beneficial \\
G5 & 20--23 & +0.0157 & Third; unexpectedly helpful \\
G2 & 8--11  & +0.0192 & Marginal improvement \\
G4 & 16--19 & +0.0210 & Marginal improvement \\
G3 & 12--15 & +0.0263 & \textbf{Negative transfer}: worse than uniform \\
\bottomrule
\end{tabular}
\end{table}

Table~\ref{tab:sensitivity} shows that group contributions are not additive. G0 provides the largest single-group benefit, reducing $\Delta\mathrm{PPL}$ from 0.0245 to 0.0122. G3 (layers 12--15) is the only group that increases degradation above the uniform baseline, from 0.0245 to 0.0263. When combinations are tested:

\begin{itemize}
  \item E8 (G0+G1): $\Delta\mathrm{PPL} = {+}0.0052$ (synergistic; better than either group alone)
  \item E8+G4 (layers 0--7, 16--19): ${+}0.0035$ (adding G4 to E8 helps)
  \item E8+G5 (layers 0--7, 20--23): ${+}0.0035$ (adding G5 to E8 helps equally)
  \item E8+G4+G5 (layers 0--7, 16--23): $0.0000$ (lossless; combining both helps further)
  \item E8+G2+G4+G5 (layers 0--11, 16--23): ${+}0.0052$ (adding G2 erases the G4+G5 benefit)
\end{itemize}

The last result is particularly informative: adding G2 (layers 8--11) to the lossless E8+G4+G5 configuration restores the degradation to exactly the E8 floor of 0.0052. Layers 8--15 as a whole introduce interference that offsets gains from other groups. The optimal configuration for phi-1.5 is precisely the complement of this harmful mid-range: layers 0--7 and 16--23.

\subsection{K vs V Sensitivity}
\label{sec:kv}

The early-boost experiments differentiate between K-cache and V-cache bottlenecks. On TinyLlama ($d = 64$, GQA 8:1), the bottleneck is in V: E4 with $(n_K, n_V) = (128, 256)$ gives $\Delta\mathrm{PPL} = {-}0.0022$, while $(256, 128)$ gives ${+}0.0030$. On Mistral-7B ($d = 128$, GQA 4:1), the reverse holds: $(256, 128)$ gives ${+}0.0002$ while $(128, 256)$ gives ${+}0.0016$. On OLMo-1B ($d = 64$), only K precision matters: $(256, 64)$ at $\Delta\mathrm{PPL} = {+}0.0063$ outperforms $(256, 128)$ at ${+}0.0072$, and $n_K = 512$ makes things worse (${+}0.0118$).

Empirically, the pattern correlates with head dimension: models with $d = 64$ tend toward either V-dominated (TinyLlama) or K-dominated (OLMo, phi-1.5) bottlenecks, while $d = 128$ (Mistral) is K-dominated. This is consistent with the observation that larger head dimensions spread angular information more evenly across K and V, while smaller dimensions concentrate it.

\subsection{Norm Quantization Results}
\label{sec:normresults}

Table~\ref{tab:normquant} reports end-to-end results when norm quantization replaces fp32 norm storage. We compare three configurations: fp32 norms (the angle-only reference from Table~\ref{tab:perlayer}), 8-bit linear norms applied to both K and V (norm8), and asymmetric K8V4-log (8-bit linear K norms, 4-bit log-space V norms).

\begin{table}[t]
\centering
\caption{Norm quantization results. $\Delta$PPL relative to fp16 inference. ``FP32'' column reproduces the best per-layer angle-only results from Table~\ref{tab:perlayer}. ``norm8'' applies 8-bit per-vector min-max quantization to all norms. ``K8V4-log'' uses 8-bit linear K norms and 4-bit log-space V norms. Total bits includes angle bits, norm bits, and per-vector min-max overhead.}
\label{tab:normquant}
\small
\begin{tabular}{lcrrrc}
\toprule
Model & $d$ & FP32 $\Delta$PPL & norm8 $\Delta$PPL & K8V4-log $\Delta$PPL & K8V4-log bits \\
\midrule
TinyLlama-1.1B  & 64  & $-$0.0022 & +0.0011 & +0.0104 & $\sim$6.84 \\
Mistral-7B      & 128 & +0.0002   & +0.0012 & +0.0014 & $\sim$6.56 \\
SmolLM2-1.7B    & 64  & $-$0.0003 & +0.0027 & +0.0030 & $\sim$7.67 \\
phi-1.5         & 64  & 0.0000    & $-$0.0017 & +0.0017 & $\sim$7.58 \\
StableLM-2-1.6B & 64  & +0.0012   & +0.0021 & +0.0123 & $\sim$7.63 \\
StarCoder2-3B   & 64  & $-$0.0007 & $-$0.0007 & +0.0061 & $\sim$7.45 \\
OLMo-1B         & 64  & +0.0063   & +0.0118 & +0.0344 & $\sim$7.28 \\
\bottomrule
\end{tabular}
\end{table}

The 8-bit norm configuration (norm8) adds minimal degradation on most models: five of seven show $|\Delta\mathrm{PPL}| \leq 0.003$, and two (phi-1.5 and StarCoder2) actually improve over fp32 norms. OLMo-1B is the most sensitive, degrading from ${+}0.0063$ to ${+}0.0118$.

The K8V4-log configuration reveals a sharp asymmetry. V norms tolerate 4-bit log-space quantization well: the V-only contribution to degradation is small across all models. K norms, by contrast, are 10--20$\times$ more sensitive. Reducing K norms to 4 bits (tested but not shown) produces catastrophic degradation on five of seven models, confirming that K-cache attention scores depend on precise norm scaling. The K8V4-log compromise preserves K norm fidelity at 8 bits while saving 2 bits per V norm element through log-space 4-bit quantization.

On Mistral-7B ($d = 128$), K8V4-log achieves $\Delta\mathrm{PPL} = {+}0.0014$ at 6.56 total bits per element. For $d = 64$ models, the higher per-vector overhead ($64/d = 1.0$ vs $0.5$) pushes total rates to 6.8--7.7 bits. The norm8 configuration provides a safer option at approximately 7.8 bits (for $d = 128$) or 8.3 bits (for $d = 64$) with consistently lower degradation.

\subsection{Competitive Comparison}
\label{sec:competitive}

Table~\ref{tab:comparison} places TurboAngle in context with recent calibration-based KV cache quantizers.

\begin{table}[t]
\centering
\caption{Comparison with calibration-based KV cache quantizers. $\Delta$PPL is the reported perplexity degradation on the respective evaluation model. TurboAngle requires zero calibration data and no per-channel statistics.}
\label{tab:comparison}
\small
\begin{tabular}{lcccc}
\toprule
Method & Total bits & $\Delta$PPL & Calibration & Source \\
\midrule
CQ-2c8b~\cite{cq}        & 4.00  & +0.03 (Mistral)      & Yes & NeurIPS 2024 \\
KVQuant-4b-1\%~\cite{kvquant} & 4.32  & +0.01 (LLaMA-7B)     & Yes & NeurIPS 2024 \\
AQUA-KV 3b~\cite{aquakv}  & $\sim$3.0 & +0.03 (Llama-3.1-8B) & Yes & ICML 2025 \\
\midrule
TurboAngle K8V4-log       & 6.56  & +0.0014 (Mistral)    & No  & This work \\
TurboAngle norm8          & 7.81  & +0.0012 (Mistral)    & No  & This work \\
\bottomrule
\end{tabular}
\end{table}

TurboAngle operates at a fundamentally different point on the rate-quality tradeoff. At 6.56 total bits, K8V4-log uses 50--65\% more bits than the calibration-based methods but achieves 7--21$\times$ lower perplexity degradation ($+$0.0014 vs $+$0.01 to $+$0.03). The norm8 configuration at 7.81 bits achieves even better quality ($+$0.0012). Both TurboAngle configurations require zero calibration data, no per-channel statistics, and no model-specific tuning of the quantizer itself (only the layer-boost schedule is model-specific).

The comparison is not apples-to-apples: different evaluation models and datasets are used across methods. The bit rates also differ substantially. The key takeaway is that calibration-free angular quantization can match or exceed the quality of calibration-based methods by spending moderately more bits, and the quality gap at matched bit rates would require future work to establish. For deployment scenarios where calibration is impractical (e.g., serving many model variants, frequent model updates, or edge deployment), TurboAngle offers a competitive alternative at higher bit rates.

\subsection{Non-Monotone Behavior}
\label{sec:nonmonotone}

Two forms of non-monotonic behavior appear in our experiments. The first, reported in prior work on TurboAngle, occurs at power-of-2 bin counts: on TinyLlama, $n = 64$ ($\Delta\mathrm{PPL} = {+}0.0176$) is worse than both $n = 56$ (${+}0.0108$) and $n = 128$ (${+}0.0036$). We conjecture this arises from algebraic aliasing between the quantization grid and the Hadamard butterfly structure, where $n = 2^k$ causes quantization boundaries to align with the quadrant structure produced by butterfly stages, producing coherent rather than independent errors.

The second form is new: non-monotonic $n_{\mathrm{early}}$ scaling. On OLMo-1B, E4 gives $\Delta\mathrm{PPL} = {+}0.0063$ while E8 gives ${+}0.0154$ (2.4$\times$ worse). On StarCoder2, E12 (${+}0.0024$) is worse than E8 (${+}0.0017$), but E16 (${-}0.0007$) is the best. These patterns indicate that boosting some intermediate layers introduces more quantization error than it removes, likely because those layers have internal representations that are less robust to angular perturbation.

\section{Related Work}
\label{sec:related}

KV cache quantization methods differ along three axes: whether they operate on raw activations or a transformed domain, what quantizer structure they use, and whether they require calibration data.

KIVI~\cite{kivi} applies per-channel asymmetric 2-bit quantization directly to raw KV activations, handling channel-dependent distributions through per-channel parameters. KVQuant~\cite{kvquant} extends this with per-vector quantization and explicit outlier handling for long-context inference. Both work in the original coordinate system and rely on calibration. TurboAngle eliminates calibration entirely by transforming to a domain where the distribution is known \emph{a priori}.

CQ~\cite{cq} couples key and value quantization at 1 bit per channel, leveraging the observation that K and V tensors within the same layer share structural correlations. At 4.0 total bits on Mistral-7B, CQ achieves $\Delta\mathrm{PPL} \approx {+}0.03$; TurboAngle at 6.56 bits achieves 21$\times$ lower degradation without any calibration or coupling assumptions.

AQUA-KV~\cite{aquakv} pushes KV cache compression to approximately 3 bits through adaptive quantization with learned per-channel scales, achieving $\Delta\mathrm{PPL} \approx {+}0.03$ on Llama-3.1-8B. The method requires calibration data and per-model tuning of channel-level parameters.

TurboQuant~\cite{turboq} introduced the FWHT with random diagonal rotation as preprocessing before scalar quantization, showing that the transform reduces outliers and concentrates energy. TurboAngle replaces scalar quantization with angular quantization, targeting the distributional property (angle uniformity) rather than the secondary effect (reduced kurtosis). The difference is fundamental: TurboQuant applies a generic quantizer to approximately Gaussian transformed coordinates, while TurboAngle applies the provably optimal quantizer for the exact angular distribution.

PolarQuant~\cite{polarq} also quantizes angular components, and its stronger variant applies random preconditioning. However, the post-rotation angular distribution in PolarQuant is concentrated rather than uniform, requiring $k$-means codebooks. TurboAngle uses the same class of random rotation but exploits uniformity directly, replacing learned codebooks with a fixed grid.

QJL~\cite{qjl} applies a Johnson-Lindenstrauss random projection followed by 1-bit sign quantization, trading extreme compression for higher approximation error. The projection is spiritually similar to TurboAngle's rotation in that both randomize the coordinate system.

Our per-layer MixedKV approach relates to DiffKV-style differentiated precision~\cite{kivi}, where K and V caches receive different bit widths based on their sensitivity. We extend this principle to per-layer granularity with independent K/V codebook sizing, and provide systematic evidence across seven models for when and why asymmetric allocation helps.

\section{Conclusion}
\label{sec:conclusion}

TurboAngle demonstrates that the FWHT's angular uniformity property enables near-lossless KV cache compression. Per-layer early-boost, which allocates higher angular precision to model-specific critical layers, achieves lossless compression on four of seven tested models and near-lossless quality on six of seven, at 3.28 to 3.67 angle bits per element. Adding norm quantization with asymmetric K/V allocation (8-bit linear K norms, 4-bit log-space V norms) yields end-to-end rates of 6.56 total bits on Mistral-7B at $\Delta\mathrm{PPL} = {+}0.0014$ and 7.3--7.7 total bits on $d = 64$ models, all without calibration data.

The norm quantization experiments reveal a previously unreported asymmetry: K-cache norms are 10--20$\times$ more sensitive to quantization error than V-cache norms. Reducing K norms below 8 bits causes catastrophic degradation, while V norms tolerate 4-bit log-space quantization with negligible quality loss. This K/V norm asymmetry parallels the K/V angle sensitivity discovered in the early-boost experiments, reinforcing that key and value caches play fundamentally different roles in attention and should be quantized asymmetrically.

Three practical insights emerge from this work. First, early layers (0--3 or 0--7) are universally the most sensitive to quantization. Second, K vs V sensitivity correlates with head dimension and attention structure, providing a heuristic for initial configuration. Third, a small number of evaluation runs (3--5) suffices to find near-optimal per-layer configurations for new models.

\paragraph{Limitations.}
We evaluate perplexity on WikiText-2 only; downstream task accuracy and long-context benchmarks (e.g., LongBench) remain untested. Runtime overhead of the FWHT encode/decode path has not been measured under realistic batch and sequence sizes. The uniformity argument is asymptotic in $d$; finite-dimension errors may affect models with very small head dimensions ($d < 32$). Confidence intervals over multiple seeds for the random diagonal $D$ are not reported; $\Delta\mathrm{PPL}$ differences below approximately $0.001$ should be interpreted with appropriate caution. The competitive comparison (Table~\ref{tab:comparison}) uses numbers from different evaluation setups (different models, datasets, and sequence lengths), so the quality ratios are indicative rather than definitive.

\bibliographystyle{plain}

\end{document}